\documentclass[manuscript]{acmart}
\AtBeginDocument{%
  }

\setcopyright{acmlicensed}
\copyrightyear{2018}
\acmYear{2018}
\acmDOI{XXXXXXX.XXXXXXX}
\acmConference[Conference acronym 'XX]{Make sure to enter the correct
  conference title from your rights confirmation email}{June 03--05,
  2018}{Woodstock, NY}
\acmISBN{978-1-4503-XXXX-X/2018/06}

\begin{document}

\title[Multilingual HateSpeech]{Multilingual Hate Speech Detection and Counterspeech Generation: A Comprehensive Survey and Practical Guide}

\author{Zahra Safdari Fesaghandis}
\affiliation{%
 \institution{Bilkent University}
  \city{Ankara}
  \country{Turkey}
}
\email{zahrasafdari8181@gmail.com}

\author{Suman Kalyan Maity}
\email{smaity@mst.edu}
\affiliation{%
  \institution{Missouri University of Science and Technology}
  \city{Rolla}
  \state{Missouri}
  \country{USA}
}







\renewcommand{\shortauthors}{Safdari et al.}


\begin{abstract}
Combating online hate speech in multilingual settings requires approaches that go beyond English-centric models and capture the cultural and linguistic diversity of global online discourse. This paper presents a comprehensive survey and practical guide to multilingual hate speech detection and counterspeech generation, integrating recent advances in natural language processing. We analyze why monolingual systems often fail in non-English and code-mixed contexts, missing implicit hate and culturally specific expressions. To address these challenges, we outline a structured three-phase framework—task design, data curation, and evaluation—drawing on state-of-the-art datasets, models, and metrics. The survey consolidates progress in multilingual resources and techniques while highlighting persistent obstacles, including data scarcity in low-resource languages, fairness and bias in system development, and the need for multimodal solutions. By bridging technical progress with ethical and cultural considerations, we provide researchers, practitioners, and policymakers with scalable guidelines for building context-aware, inclusive systems. Our roadmap contributes to advancing online safety through fairer, more effective detection and counterspeech generation across diverse linguistic environments.
\end{abstract}



\received{13 September 2025}
\received[revised]{}
\received[accepted]{}

\maketitle
\section{Introduction}
With the rise of social media, global communication has been transformed, enabling unprecedented user engagement while simultaneously amplifying harmful content such as hate speech \citep{fortuna2018survey}. Hate speech—defined as language disparaging individuals or groups based on attributes such as ethnicity, religion, or gender—poses significant threats to online safety and democratic discourse \citep{davidson2017automated,ousidhoum2019multilingual}. Beyond its immediate harm to targeted communities, hate speech also fosters polarization, undermines trust in institutions, and fuels offline violence, highlighting its broad societal impact \citep{mathew2021hatexplain,vidgen2020directions}. This problem is particularly acute in multilingual environments, where models primarily trained on English often struggle to generalize to linguistically and culturally diverse settings \citep{aluru2020deep,muhammad2025afrihate}. The global nature of online platforms means that harmful speech is not confined to one language or region. On platforms such as Twitter, Facebook, TikTok, and YouTube, users interact in multiple languages, often combining dialects, scripts, and informal variations within a single post \citep{Arora2021DetectingHC}. This reality creates challenges for traditional moderation pipelines, which are frequently optimized for English or other high-resource languages. Research demonstrates that hate speech in low-resource languages is disproportionately overlooked due to data scarcity, code-mixing (e.g., Hinglish, Arabizi), and culturally specific expressions that evade English-centric detection models \citep{yadav2023hate,chavinda2025dual}. Even translation-based solutions risk losing socio-cultural nuance, leading to misinterpretation, unfair moderation, or over-censorship of marginalized voices \citep{chan2024hate,rottger2022multilingual}. These issues exacerbate systemic inequalities, allowing harmful narratives against vulnerable groups in non-English contexts to remain undetected or inadequately mitigated.

In addition to linguistic diversity, the multimodal nature of online hate further complicates detection. Hate speech is increasingly conveyed not only through text but also via memes, emojis, videos, and other multimodal artifacts \citep{Lippe2020AMF,botelho2021deciphering}. Detecting such content requires integrating vision, audio, and textual cues, a challenge that remains underexplored in multilingual contexts. Furthermore, implicit hate—such as sarcasm, coded speech, dog whistles, and stereotypes—is especially difficult to capture with automated systems and often escapes keyword-based or literal interpretation approaches \citep{elsherief2021latent,talat2016hateful}. These complexities highlight the urgent need for more robust, culturally grounded, and multimodal detection strategies. Counterspeech has emerged as a promising alternative to content removal, defined as non-aggressive, informed responses that challenge hate while preserving freedom of expression \citep{chung2019conan,mathew2019thou,fanton2021human}. By fostering dialogue rather than silencing voices, counterspeech can de-escalate conflict and promote healthier online interactions. However, counterspeech research faces its own challenges: datasets are often small, fragmented, and culturally uneven; taxonomies vary across contexts; and evaluation remains inconsistent, with automated metrics like BLEU underestimating persuasiveness and cultural appropriateness \citep{bonaldi2025first,das2024low,sahoo2024indicconan}. Effective counterspeech further requires balancing persuasiveness, empathy, and cultural sensitivity—qualities that are difficult to model computationally and even harder to evaluate reliably across diverse languages.

Against this backdrop, this survey integrates recent advances in multilingual NLP and provides both a comprehensive background and a practical roadmap for hate speech detection and counterspeech generation across diverse linguistic and cultural contexts. Specifically, we outline a three-phase framework: (i) \textbf{task design}, focusing on classification, generation, and cross-lingual transfer strategies; (ii) \textbf{data curation}, highlighting high-quality multilingual datasets and annotation practices; and (iii) \textbf{evaluation}, combining quantitative metrics, qualitative human assessments, and fairness measures. Unlike prior surveys that focus predominantly on English or high-resource languages, we emphasize inclusivity, fairness, and cultural awareness as central design principles. Finally, we highlight persistent open challenges—including data scarcity in low-resource languages, translation inaccuracies, cultural subjectivity, and the need for multimodal approaches—and propose future directions for addressing them \citep{bui2024multi3hate,bonaldi2024nlp,muhammad2025afrihate}. By doing so, we aim to equip researchers, practitioners, and policymakers with the tools needed to foster safer, more inclusive digital spaces worldwide. Beyond technical contributions, we argue that meaningful progress in this domain requires interdisciplinary collaboration across computer science, linguistics, and social sciences, as well as partnerships with local communities and policymakers to ensure that solutions are not only scalable but also contextually appropriate and ethically responsible.

\section{Background}
\label{sec:Background}
The rise of social media has amplified user-generated content, including hate speech, posing significant challenges to maintaining safe online environments. Traditional moderation approaches, such as content removal, often fall short, prompting growing interest in counterspeech as a constructive alternative \citep{chung2019conan}. This section outlines key concepts in multilingual NLP research on hate speech and counterspeech, providing definitions, taxonomies, and relevant task formulations to contextualize these challenges.

\subsection{What Is Hate Speech?}
\label{subsec:defining_hate_speech}

Hate speech refers to language that humiliates or attacks individuals or groups based on protected attributes (e.g., race, ethnicity, religion, gender, and/or sexual orientation) \citep{fortuna2018survey, aluru2020deep, rottger2022multilingual}. However, what constitutes hate speech may change with context, and even vary with legal, cultural and linguistic variation. For example, \citet{davidson2017automated} note that while hate speech is protected under free speech laws in the United States, many countries define it narrowly as targeting minorities in ways that incite violence or social unrest. These differences present difficulties for automated detection systems, especially in multilingual settings, where the confluence of cultural sensitivities and linguistic heterogeneity contributes to definitional complexities \citep{ousidhoum2019multilingual, muhammad2025afrihate}.

In multilingual settings, hate speech identification has further challenges, such as the scarcity of data for low-resource languages and the need for the development of efficient cross-lingual transfer approaches \citep{aluru2020deep}. \citet{aluru2020deep} illustrate that the majority of hate speech datasets are in English and that models perform poorly for languages such as Arabic, Indonesian, or African languages such as Amharic and Swahili \citep{muhammad2025afrihate}. While data translation to English can be a good strategy for detection, it may result in loss of the socio-cultural context \citep{chan2024hate}. \citet{rottger2022multilingual} emphasize the contribution of native-speaker annotations in capturing linguistic and cultural relevance, which is evident in their multilingual \textit{HateCheck} suite for ten languages. Similarly, \citet{muhammad2025afrihate} promote local community involvement in defining hate speech in African languages to address issues like class imbalance and language identification. These insights highlight the importance of strong, culturally aware definitions and varied datasets for improving the detection of multilingual hate speech.

\textbf{Taxonomies of Hate Speech:} Hate speech taxonomies provide structured frameworks to classify content by target, type, or intent, enabling fine-grained detection critical for multilingual applications \citep{fortuna2018survey}. \citet{ousidhoum2019multilingual} propose a multi-aspect annotation schema for English, French, and Arabic tweets, categorizing hate speech by directness (direct or indirect), offensiveness (e.g., hateful, abusive), discriminatory attribute (e.g., ethnicity, gender), target group, and sentiment. This framework transcends binary classification, capturing the complexity of hate speech across languages. Similarly, \citet{siddiqui2024fine} classify hate speech in English, Urdu, and Sindhi into five fine-grained categories—Disability, Gender, Nationality, Race, and Religion—emphasizing the need for nuanced taxonomies in low-resource languages.

\citet{davidson2017automated} and \citet{mathew2021hatexplain} adopt a three-class taxonomy (hate speech, offensive language, or neither), with \citet{mathew2021hatexplain} extending it in HateXplain to include target community annotations (e.g., ethnicity, religion) and explanatory rationales. \citet{basile2019semeval} focus on hate speech against immigrants and women in English and Spanish, classifying it by presence, aggressiveness (aggressive or non-aggressive), and target (individual or group). \citet{zampieri2019predicting} propose a hierarchical taxonomy for offensive language, including hate speech, with layers for offensiveness (offensive or not), targeting (targeted or untargeted), and target type (e.g., individual or group, gender, race). \citet{muhammad2025afrihate} label hate speech in African languages by discriminatory attributes (e.g., ethnicity, politics), offering a foundational categorization adaptable to low-resource settings.

These taxonomies enhance detection by capturing diverse manifestations of hate speech. However, their application in multilingual contexts requires addressing linguistic and cultural variations, as noted by \citet{ousidhoum2019multilingual, basile2019semeval}. Extending these frameworks to under-resourced languages remains a critical research direction.

\subsection{What Is Counterspeech?}
\label{subsec:defining_counter_speech}

Counterspeech, often termed counter-narratives, comprises non-aggressive, informed textual responses designed to mitigate hate speech by offering alternative viewpoints, factual rebuttals, or de-escalatory arguments \citep{chung2019conan, fanton2021human, bengoetxea2024basque, sahoo2024indicconan, bennie2025codeofconduct}. Below, we detail specific approaches, examples, and challenges in multilingual counterspeech, drawing on recent research.

\citet{chung2019conan} describe counter-narratives as expert-crafted responses that challenge hate speech without censorship, developed through niche-sourcing with Non-Governmental Organization (NGO) operators in English, French, and Italian. Similarly, \citet{fanton2021human} emphasize respectful responses that foster healthier online discourse, while \citet{sahoo2024indicconan} define counter-narratives as fact-based rebuttals to stereotypes in Hindi and Indian English. \citet{bengoetxea2024basque} focus on evidence-based responses to Islamophobia in Basque and Spanish, and \citet{bennie2025codeofconduct} highlight context-aware counterspeech in low-resource languages like Basque. \citet{mathew2019thou} further refine the definition, specifying counterspeech as direct comments, distinct from replies to other comments, targeting hateful or harmful video content.

Multilingual counterspeech faces challenges, including data scarcity, translation fidelity, and cultural alignment. \citet{chung2019conan} address this by creating parallel corpora through translating French and Italian counter-narratives into English, enabling cross-lingual research. \citet{bengoetxea2024basque} use machine translation with professional post-editing to develop CONAN-EUS, noting that linguistic similarity influences transfer effectiveness. \citet{sahoo2024indicconan} tackle low-resource Indic languages using the NLLB translator with human validation to ensure grammatical and cultural accuracy. \citet{bennie2025codeofconduct} demonstrate robust counterspeech generation in Basque, leveraging multilingual datasets and optimization techniques. These efforts highlight the need for culturally sensitive, high-quality datasets and models to support counterspeech across diverse linguistic contexts.


\textbf{Counterspeech Taxonomies}: Counterspeech taxonomies categorize responses by type, strategy, or target, facilitating targeted interventions against hate speech \citep{chung2019conan}. \citet{chung2019conan} introduce a comprehensive taxonomy, including strategies like presenting facts, pointing out hypocrisy, warning of consequences, affiliation, positive tone, negative tone, humor, and counter-questions, applied across English, French, and Italian. \citet{chung2021multilingual} refine this into five primary categories—facts, denouncing, questioning, hypocrisy, and humor—alongside non-counter-narrative types (support, unrelated), achieving strong classification results in multilingual settings. \citet{sahoo2024indicconan} propose a taxonomy for Hindi and Indian English, encompassing consequences, denouncing, facts, contradiction, counter-questions, and positive responses, tailored to the Indian context.

\citet{das2024low} introduce a strategy-based taxonomy for Bengali and Hindi, including warnings, shaming and labeling, empathy, pointing out hypocrisy, affiliation, and humor. Their findings indicate that monolingual training outperforms cross-lingual models in these languages due to significant linguistic diversity. \citet{fanton2021human} implicitly classify counter-narratives by target groups (e.g., Women, POC, LGBT+), enabling alignment with hate speech targets, though their framework lacks explicit type-based categorization. These taxonomies enhance counterspeech generation by capturing diverse strategies, but their application in multilingual contexts requires addressing linguistic and cultural variations, as noted by \citet{chung2021multilingual, das2024low}. Extending these frameworks to under-represented languages remains a pressing research challenge.
\begin{table}[t]
\centering
\begin{tabular}{@{}p{1.2cm}p{6cm}@{}}
\toprule
\textbf{Text} & "A nigr*** too dumb to f*** has a scant chance of understanding anything beyond the size of a d***." \\
\textbf{Label} & Hate \\
\textbf{Targets} & Women, African \\
\bottomrule
\end{tabular}
\caption{Example of hate speech from the HateXplain \cite{mathew2021hatexplain} dataset, illustrating targeted derogatory language for research purposes.}
\label{tab:HS-example}
\end{table}

\begin{table}[t]
\centering
\begin{tabular}{@{}p{2cm}p{6cm}@{}}
\toprule
\textbf{Hate Speech} & "Muslims conceived the slave trade." \\
\textbf{Counterspeech} & "Slavery long predated Islam; they inherited slavery and proceeded to improve conditions. Way ahead of the rest of the world." \\
\textbf{Counter Type} & Presentation of facts \\
\bottomrule
\end{tabular}
\caption{Example of a counter-narrative from the CONAN \cite{chung2019conan} dataset, illustrating the ``presenting facts'' strategy.}
\label{tab:conan-example}
\end{table}

\section{Design Your Task}
\label{sec:Task-Design}

Designing tasks for multilingual hate speech detection and counterspeech is key to tackling online hate across cultures. This section outlines strategies for structuring classification, generation, and cross-lingual tasks with a focus on scalability, fairness, and cultural sensitivity. Well-designed tasks not only provide a foundation for benchmarking but also reveal the trade-offs between robustness, adaptability, and ethical considerations in real-world applications. By organizing tasks around detection and response, researchers can create pipelines that align linguistic processing with broader societal goals..

\subsection{Tasks for Multilingual Hate Speech Detection}
\label{subsec:hate_speech_tasks}

Effective hate speech detection requires tasks that navigate linguistic diversity and cultural nuances. Here, we discuss classification tasks, cross-lingual methods, and best practices, critically assessing their strengths and limitations with key techniques summarized in Table~\ref{tab:hate_speech_techniques}. Task design is not only a technical challenge but also a cultural one, as the formulation of labels, categories, and evaluation criteria directly influences how systems perform across different linguistic communities.
\begin{table}[!ht]
\centering
\resizebox{.95\textwidth}{!}{
\begin{tabular}{@{}p{3.5cm}p{6cm}p{4cm}p{2cm}@{}}
\toprule
\textbf{Technique} & \textbf{Languages} & \textbf{Relevance to Task Design} & \textbf{References} \\
\midrule
BERT-based Classification & English, Arabic, German, Indonesian, Italian, Polish, Portuguese, Spanish, French, Hindi, Urdu, Sindhi, Hinglish & Binary, multi-label tasks for diverse languages & \cite{aluru2020deep, siddiqui2024fine, shukla2024multilingual} \\
Multi-aspect Classification & English, French, Arabic, Devanagari languages & Fine-grained tasks for directness, target, sentiment & \cite{ousidhoum2019multilingual, thapa2025natural} \\
Multi-label with Explainability & English, Urdu, Sindhi & Transparent multi-label tasks & \cite{mathew2021hatexplain, siddiqui2024fine} \\
Multimodal Classification & English, German, Spanish, Hindi, Mandarin, Hinglish & Text-image, real-time tasks & \cite{bui2024multi3hate, shukla2024multilingual} \\
Zero-shot Learning & English, Italian, Spanish, German, Arabic, Greek, Turkish & Cross-lingual tasks without target data & \cite{nozza2021exposing, zia2022improving, bigoulaeva2021cross} \\
Few-shot Learning & English, Norwegian, Arabic, Spanish, German, Italian, French, Portuguese & Low-resource tasks with minimal data & \cite{mozafari2022cross, hashmi2025metalinguist} \\
Transfer Learning & English, German, Hindi, Chinese & Scalable cross-lingual tasks & \cite{roy2021leveraging, liu2023cross, bigoulaeva2021cross} \\
Multilingual Embeddings & English, Spanish, German, Russian, Turkish, Croatian, Albanian & Contextual cross-lingual representations & \cite{conneau2019unsupervised, bojkovsky2019stufiit, de2022unsupervised} \\
Multi-task Learning & English, Hindi, German, Spanish, Italian & Generalizable tasks across sub-tasks & \cite{mishra2021exploring, montariol2022multilingual} \\
Semi-supervised GANs & English, German, Hindi & Data augmentation for low-resource tasks & \cite{mnassri2024multilingual} \\
Human-in-the-loop & English, Spanish, German, French & Robust, adaptive task design & \cite{vitiugin2021efficient, kotarcic2022human} \\
Fairness Evaluation & English, Italian, Polish, Portuguese, Spanish, Indonesian, German, French, Arabic & Non-discriminatory task design & \cite{huang2020multilingual, ousidhoum2020comparative} \\
Ensemble Methods & English, Bengali, Indonesian, Italian, Spanish & Balanced, robust classification & \cite{mahajan2024ensmulhatecyb} \\
Lightweight Models & Arabic, English, Turkish, Hindi, Italian, Spanish, Indonesian, German, Portuguese, Danish, Malay, French & Scalable tasks for diverse languages & \cite{kousar2024mlhs} \\
Federated Learning & Hindi, Tamil, Telugu, Kannada, Malayalam, Bengali, Marathi, Bhojpuri, Gujarati, Haryanvi, Odia, Punjabi, English & Privacy-preserving, generalizable tasks & \cite{singh2024generalizable} \\
Feature Engineering & English, Slovene, Dutch, Hindi, German & Enhanced cue detection & \cite{markov2021exploring, mishra2021exploring} \\
Synthetic Data & English, German, Greek, Italian, Hindi & Addressing data scarcity & \cite{korre2024challenges, mnassri2024multilingual} \\
Topic Modeling for Bias & English, French, German, Arabic, Italian, Portuguese, Indonesian & Balanced dataset design & \cite{ousidhoum2020comparative} \\
\bottomrule
\end{tabular}}
\caption{Summary of Techniques for Designing Multilingual Hate Speech Detection Tasks in Diverse Linguistic Contexts.}
\label{tab:hate_speech_techniques}
\end{table}

\subsubsection{Classification Tasks}
Hate speech classification is commonly framed as binary (hate vs. non-hate) or multi-class (e.g., racism, sexism) tasks. \citet{aluru2020deep} employ BERT-based models for binary classification across nine languages, demonstrating scalability in multilingual settings. Multi-task learning offers a nuanced alternative, as shown by \citet{ousidhoum2019multilingual}, who use a unified model to classify five hate speech aspects (e.g., directness, target group) in English, French, and Arabic, enabling simultaneous learning of correlated dimensions. These examples highlight the trade-off between simplicity and nuance: while binary classification is easier to scale, multi-task and multi-class approaches capture richer sociolinguistic variation at the expense of data requirements and annotation complexity.

Multi-label classification addresses overlapping hate categories. \citet{mathew2021hatexplain} implement multi-label classification with explainability in English, labeling toxicity and stereotype dimensions. \citet{siddiqui2024fine} apply Transformer-based multi-label classification in English, Urdu, and Sindhi, using LIME \citep{ribeiro2016should} for transparency to support clear decision-making. \citet{shukla2024multilingual} propose a multilingual BERT-based framework for Hinglish, integrating multimodal data to detect hate across diverse content formats, critical for real-time social media applications. \citet{gertner2019mitre} extend classification to multi-aspect tasks in English and Spanish, identifying hate presence, target, and aggression, formulated as a five-class problem or separate predictions. These task formulations show how explainability and multi-dimensional labels improve trustworthiness and interpretability, but they also require fine-grained taxonomies and more careful quality control in multilingual contexts.

Fine-grained classification tasks target specific hate speech aspects or domains. \citet{thapa2025natural} define a shared task for Devanagari-script languages (Hindi, Nepali, Marathi, Sanskrit, Bhojpuri), including binary hate detection and target classification leveraging Transformer models and multilingual embeddings. \citet{de2018multilingual} analyze cross-domain hate speech (e.g., jihadism, extremism, sexism, racism) across English, Arabic, German, Dutch, and French, proposing multi-label tasks for diverse hate types. \citet{mahajan2024ensmulhatecyb} introduce EnsMulHateCyb, an ensemble model for binary classification across English, Bengali, Indonesian, Italian, and Spanish, combining offensive and hate speech categories to address cyberbullying. \citet{singh2024generalizable} propose MultiFED, a federated learning approach for binary classification in low-resource Indian languages (e.g., Hindi, Tamil, Telugu), using fair client selection to enhance generalizability. These methods illustrate how task design is closely tied to deployment needs: fine-grained labeling supports detailed moderation, while federated approaches enable scalable deployment without compromising privacy or fairness.

Code-mixed languages present unique challenges. \citet{biradar2021hate} propose TIF-DNN for Hinglish, using translation and transliteration to convert code-mixed data to monolingual Hindi before classification, improving performance over direct processing. \citet{yadav2023hate} evaluate deep learning models (e.g., CNN-BiLSTM with word2vec) for Hinglish, designing tasks to handle code-mixing and non-standard writing. Multimodal tasks incorporate non-textual data, as in \citet{bui2024multi3hate}, which uses vision-language models for text-image hate detection across English, German, Spanish, Hindi, and Mandarin, essential for social media platforms. \citet{muhammad2025afrihate} provide datasets for 15 African languages, enabling binary and multi-class tasks in low-resource contexts like Amharic and Swahili. \citet{ghosh2025hate} analyze Transformer models for low-resource Indian languages (Hindi, Marathi, Bangla, Assamese, Bodo), designing binary tasks with cross-lingual transfer to address data scarcity. \citet{kousar2024mlhs} propose MLHS-CGCapNet, a lightweight model for binary classification across 12 languages, leveraging convolutional and capsule networks for linguistic diversity. \citet{hashmi2025metalinguist} design binary tasks for English and Norwegian, using meta-learning for low-resource settings. \citet{mnassri2024multilingual} use semi-supervised GANs for binary classification in English, German, and Hindi, tackling labeled data scarcity. \citet{pacaldo2025leveraging} frame multilingual hate speech detection as a binary classification task for Cebuano, Tagalog, and English, using traditional ML (e.g., SVM, Random Forest) and transformers (mBERT, XLM-RoBERTa) with SMOTE for imbalance and hyperparameter tuning via Grid Search. \citet{usman2025large} frame binary hate speech detection for English, Spanish, and Urdu using ML (SVM, RF), DL (BiLSTM, CNN), TL (BERT, XLM-RoBERTa), and LLMs (GPT-3.5-turbo), with translation-based pipelines for standardization. \citet{mnassri2025rag} frame binary hate speech detection using HS-RAG and HS-MemRAG with Meta-LLaMA-3-8B, enhancing performance in imbalanced Arabic datasets via retrieval. \citet{ahmad2025ua} design binary and multi-class (Direct, Disguised, Sarcastic, Exclusionary) hate speech tasks for Arabic and Urdu using XLM-RoBERTa. Collectively, these works underline that task formulations must be adaptable: what works in one linguistic setting (e.g., Hinglish code-mixing) may not directly apply to others (e.g., Arabic sarcasm), and designing robust pipelines often requires a combination of translation, augmentation, and culturally aware annotation.

\subsubsection{Cross-Lingual Methods}
Cross-lingual methods extend hate speech detection to low-resource languages, reducing dependence on labeled data \citep{conneau2019unsupervised}. Zero-shot learning enables models trained on high-resource languages to predict hate in untrained ones. \citet{nozza2021exposing} highlight challenges in zero-shot transfer across English, Italian, and Spanish, noting misinterpretations of language-specific nuances. \citet{zia2022improving} use pseudo-label fine-tuning to enhance zero-shot performance across six non-English languages. \citet{bigoulaeva2021cross} employ cross-lingual transfer from English to German, using bilingual word embeddings and neural classifiers (CNNs, BiLSTMs) in a zero-shot setup, leveraging unlabeled target data via bootstrapping. \citet{montariol2022multilingual} improve zero-shot transfer by training on auxiliary tasks like sentiment analysis and named entity recognition, using m-BERT and XLM-R across English, Spanish, and Italian. For low-resource Philippine languages, \citet{pacaldo2025leveraging} leverage mBERT's cross-lingual capabilities, achieving high generalization through fine-tuning on combined datasets. \citet{usman2025large} apply joint multilingual and translation approaches (via Google Translate) for cross-lingual transfer, boosting LLM performance in low-resource Urdu. \citet{yoo2025adaptive} extend PMF with meta-learners (e.g., Random Forest) across English, Korean, Chinese, and Portuguese, leveraging multilingual embeddings for cross-lingual robustness. These approaches illustrate that while cross-lingual transfer holds promise for equitable coverage, performance is often uneven, with high-resource languages dominating and subtle sociocultural signals lost in transfer.

Few-shot learning leverages limited target-language data. \citet{mozafari2022cross} apply meta-learning across eight languages, using MAML and Proto-MAML to outperform transfer learning baselines. \citet{hashmi2025metalinguist} propose meta-learning for English and Norwegian, supporting zero-shot and few-shot scenarios with transformers like Nor-BERT. Transfer learning scales detection, with \citet{roy2021leveraging} using pre-trained Transformers for English, German, and Hindi, and \citet{liu2023cross} applying contrastive learning for English, German, and Chinese. \citet{bojkovsky2019stufiit} explore multilingual embeddings (MUSE, ELMo) with adversarial learning for English and Spanish, emphasizing contextual embeddings in cross-lingual tasks. These works show that task design for cross-lingual transfer cannot rely solely on technical alignment; instead, success depends on how well training data reflects real-world usage in both source and target languages.

\textbf{Best Practices:} Effective task design for hate speech detection prioritizes robustness, scalability, and fairness. Several strategies have emerged as best practices for multilingual hate speech detection. To address \textit{data scarcity}, researchers employ translation-based methods and GAN-based augmentation, though semantic fidelity remains a challenge \citep{korre2024challenges, mnassri2024multilingual}. Beyond data augmentation, \textit{feature engineering} has proven effective: stylometric and emotion-based features outperform traditional n-grams for English, Slovene, and Dutch \citep{markov2021exploring}, while word2vec with CNN-BiLSTM enhances contextual representation for code-mixed Hinglish \citep{yadav2023hate}. \textit{Ensemble methods} such as BiLSTM, BiGRU, and CNN-LSTM combined with GloVe embeddings balance model strengths, achieving robust performance across languages \citep{mahajan2024ensmulhatecyb}. For efficiency, \textit{lightweight models} like convolutional and capsule networks, exemplified by MLHS-CGCapNet, handle linguistic diversity effectively across 12 languages \citep{kousar2024mlhs}. \textit{Transformer-based models} remain the cornerstone for scalability, with mBERT and XLM-RoBERTa widely used \citep{roy2021leveraging, ghosh2025hate}, and MuRIL-BERT particularly suited for low-resource Indian languages. Human oversight is also essential: \textit{human-in-the-loop pipelines} integrate annotator feedback to enhance adaptability, as shown by attention networks for English and Spanish \citep{vitiugin2021efficient} and BERT-based moderation pipelines for German and French \citep{kotarcic2022human}. Ensuring reliability across contexts requires \textit{robust testing}, including functional validation across multiple languages and data augmentation methods like SMOTE to address imbalance \citep{chawla2002smote, rottger2022multilingual}. Transparency is equally critical: \textit{explainability techniques} such as LIME improve interpretability for multilingual tasks \citep{siddiqui2024fine}. In addition, \textit{fairness evaluation} has become a priority, with metrics like FNED and FPED used to detect demographic bias across Italian, Polish, Portuguese, and Spanish \citep{huang2020multilingual, ousidhoum2020comparative}, and label-agnostic topic models applied to mitigate selection bias. Hybrid architectures also show promise: \textit{hybrid ML--Transformer approaches}, such as combining traditional classifiers with mBERT, improve performance for low-resource Cebuano with techniques like SMOTE and Chi-Square filtering \citep{pacaldo2025leveraging}. Similarly, \textit{LLM integration} has enabled multilingual classification with GPT-3.5-turbo in low-resource Urdu, aided by translation-based data harmonization \citep{usman2025large}. Advanced ensembles further enhance performance, as \textit{BERT ensembles with PMF} improve accuracy for low-resource Korean by combining mBERT and KoBERT adaptively \citep{yoo2025adaptive}. Finally, retrieval-augmented generation techniques such as \textit{HS-MemRAG with LLaMA-3-8B} reduce redundancy and improve classification in low-resource Arabic settings \citep{mnassri2025rag}. Collectively, these practices---though resource-intensive---are critical for designing multilingual hate speech detection systems that generalize across diverse contexts while balancing technical rigor with ethical considerations.

\subsection{Tasks for Multilingual Counterspeech}
\label{subsec:counterspeech_tasks}

Multilingual counterspeech mitigates hate by delivering constructive, culturally sensitive responses \citep{chung2019conan}. Effective task design is crucial for classifying, generating, and evaluating counterspeech, particularly in low-resource languages like Basque and Hindi. Unlike detection tasks, which focus on identifying and categorizing harmful content, counterspeech requires designing responses that balance persuasion, empathy, and cultural fit. The complexity of this challenge lies in creating systems that both generalize across languages and remain sensitive to community-specific contexts. This subsection outlines tasks, cross-lingual methods, and best practices, with summarized key techniques in Table~\ref{tab:counterspeech_techniques}.
\begin{table}[!ht]
\centering
\resizebox{.95\textwidth}{!}{
\begin{tabular}{@{}p{3.5cm}p{3.5cm}p{4cm}p{4cm}@{}}
\toprule
\textbf{Task/Technique} & \textbf{Languages} & \textbf{Relevance to Task Design} & \textbf{References} \\
\midrule
\textbf{Classification} & & & \\
Nichesourcing Annotation & English, French, Italian & Classifying counter-narrative types (e.g., facts, humor) & \citet{chung2019conan, chung2021multilingual} \\
\textbf{Generation} & & & \\
Targeted Fine-Tuning & English, Spanish & High-quality, argumentative response generation & \citet{furman2023high, furman2022parsimonious} \\
LLM Optimization & Basque, English, Italian, Spanish & Aligned, context-aware generation for low-resource languages & \citet{wadhwa2024northeastern, bennie2025codeofconduct} \\
\textbf{Evaluation} & & & \\
LLM-based Judging & Basque, English, Italian, Spanish & Assessing linguistic and contextual appropriateness & \citet{bennie2025codeofconduct} \\
Background Knowledge Integration & Basque, English, Italian, Spanish & Enhancing evaluation robustness & \citet{bonaldi2025first} \\
\textbf{Cross-Lingual Methods} & & & \\
Zero-shot Generation & English, Basque, Italian, Spanish & Generation without target language data & \citet{moscato2025mnlp} \\
Data and Model Transfer & English, Basque, Spanish & Scalable generation using multilingual datasets & \citet{bengoetxea2024basque, farhan2025hyderabadi} \\
Interlingual Transfer & Bengali, Hindi & Optimizing generation for related languages & \citet{das2024low} \\
\textbf{Best Practices} & & & \\
Human-in-the-loop Curation & Hindi, Indian English, Multilingual & High-quality datasets for regional contexts & \citet{sahoo2024indicconan, fanton2021human} \\
Knowledge-driven Approaches & English, Basque, Italian, Spanish & Improving factual accuracy and semantic similarity & \citet{marquez2025nlp, russo2025trenteam} \\
\bottomrule
\end{tabular}}
\caption{Summary of Key Techniques for Multilingual Counterspeech Task Design in Diverse Linguistic Contexts}
\label{tab:counterspeech_techniques}
\end{table}

\subsubsection{Classification, Generation, and Evaluation Tasks}
Classification tasks categorize counterspeech types for strategic interventions. \citet{chung2019conan} use nichesourcing to annotate types (e.g., facts, humor) for Islamophobia hate targets. \citet{chung2021multilingual} classify five types (e.g., denouncing, questioning), enabling multilingual applications, though expert input limits scalability. Such classification provides a taxonomy that guides generation tasks, as different hate contexts may call for distinct counterspeech strategies (e.g., factual correction versus empathetic reframing). The challenge lies in developing typologies that are both comprehensive and culturally flexible, since the perceived effectiveness of humor, questioning, or factual correction may vary across societies.

Generation tasks produce tailored counterspeech. \citet{furman2023high} demonstrate targeted fine-tuning with small, high-quality datasets in English and Spanish, enhancing argumentative quality. \citet{wang2024intent} use dual discriminators to guide Large Language Models (LLMs) for intent-aligned counterspeech, ideal for specific hate categories. \citet{hengle2024intent} employ instruction tuning and reinforcement learning for non-toxic responses, prioritizing ethical impact. \citet{wadhwa2024northeastern} align LLMs with Direct Preference Optimization for Basque, Italian, and Spanish, while \citet{bennie2025codeofconduct} use simulated annealing for context-aware generation in low-resource Basque. \citet{lyu2025hw} apply a generation-reranking pipeline, enhancing diversity in English and Italian. \citet{furman2022parsimonious} annotate argumentative elements to improve generation, supporting persuasive responses. Collectively, these works show a progression from rule-based or template-based outputs to intent-driven, preference-aligned generation that emphasizes ethical grounding and argumentative strength. Generation tasks therefore combine linguistic fluency with pragmatic goals: counterspeech must be coherent, non-toxic, persuasive, and culturally attuned all at once.

Evaluation tasks assess counterspeech quality. \citet{bennie2025codeofconduct} use LLM-based judging (JudgeLM \citep{zhu2023judgelm}) and round-robin tournaments to evaluate linguistic and contextual appropriateness in multilingual settings. \citet{bonaldi2025first} leverage the ML-MTCONAN-KN dataset to explore the effectiveness of incorporating background knowledge, enhancing evaluation robustness. These evaluation strategies illustrate how task design must move beyond surface metrics such as BLEU or ROUGE, which fail to capture pragmatic effectiveness. Instead, combining automatic evaluation with context-aware human or LLM-based judging allows for more reliable assessment of whether counterspeech is persuasive, respectful, and safe to deploy. Evaluation remains a key bottleneck: a system may achieve high fluency yet fail pragmatically, making evaluation tasks indispensable to ensuring real-world readiness.

Taken together, classification, generation, and evaluation tasks are interdependent. Classification provides structured categories of counterspeech strategies; generation produces responses aligned with those categories; and evaluation verifies their appropriateness in real-world contexts. Weakness in one stage cascades to the others—for instance, a limited classification scheme can restrict generation diversity, while inadequate evaluation metrics may misrepresent actual effectiveness. Task design for counterspeech is thus most successful when conceived as a holistic pipeline.

\subsubsection{Cross-Lingual Methods}
Cross-lingual methods enable counterspeech in low-resource languages by leveraging high-resource data and models. Zero-shot generation transfers models trained on high-resource languages to untrained ones. \citet{moscato2025mnlp} employ zero-shot generation with Mistral-7B-Instruct for Basque, Italian, and Spanish, highlighting direct generation’s effectiveness over translation. Data and model transfer utilize multilingual datasets or models. \citet{bengoetxea2024basque} use mT5 with post-edited translations for Basque and Spanish, emphasizing quality curation. \citet{farhan2025hyderabadi} fine-tune LLMs for Basque and transfer to Italian and Spanish, demonstrating scalability. These approaches reveal the tension between efficiency and authenticity: while translation-based pipelines enable rapid scaling, they may distort tone or nuance, whereas native-language generation offers better alignment with cultural norms.

Interlingual transfer optimizes performance across related languages. \citet{das2024low} explore monolingual and joint training for Bengali and Hindi, finding monolingual setups superior but interlingual transfer effective for similar languages. This shows how leveraging structural and semantic similarities between languages can reduce annotation costs without sacrificing quality. However, success in closely related languages does not guarantee effectiveness across distant ones, and cultural resonance often matters as much as linguistic similarity. Counterspeech effectiveness depends on whether the generated responses reflect not only correct grammar but also appropriate tone, rhetorical style, and cultural norms.

Cross-lingual methods thus extend coverage to underrepresented languages and reduce barriers for communities most affected by online hate. Still, they pose risks: zero-shot generation may carry biases from dominant languages, and translation may fail to preserve humor, empathy, or politeness markers critical to counterspeech effectiveness. Task design therefore requires careful calibration between maximizing reach and safeguarding contextual sensitivity.

\textbf{Best Practices:} Effective task design for counterspeech detection and generation must prioritize cultural sensitivity, persuasiveness, and ethical safeguards. One key practice is \textit{human-in-the-loop curation}, where expert annotators ensure cultural relevance. For example, \citet{sahoo2024indicconan} highlight how human validation helps address caste-based hate in Hindi and Indian English, while \citet{fanton2021human} demonstrate that expert-curated multi-target datasets improve diversity and inclusivity. Another promising direction is \textit{knowledge-driven approaches}, which leverage contextualized knowledge graphs to improve semantic accuracy and coherence. As shown by \citet{marquez2025nlp} and \citet{russo2025trenteam}, such methods enhance similarity judgments and improve passage re-ranking across English, Basque, Italian, and Spanish. \textit{Robust evaluation} frameworks are also essential, combining automated metrics like BLEU and JudgeLM with human ratings to better capture linguistic quality and persuasive impact; this approach has been shown effective in multilingual settings by \citet{moscato2025mnlp} and \citet{bennie2025codeofconduct}, particularly when background knowledge is filtered for coherence. Ethical safeguards are equally important: \textit{ethical design} practices such as instruction tuning help optimize systems for non-toxicity, preventing unintended harm during counterspeech generation \citep{hengle2024intent}. Finally, incorporating \textit{argumentative strategies} has been shown to enhance persuasiveness. As \citet{furman2022parsimonious} demonstrate, annotating argumentative elements allows counterspeech models to produce more effective, context-aware responses. Collectively, these practices provide a roadmap for building counterspeech systems that are culturally grounded, ethically responsible, and persuasive across diverse multilingual contexts.

\section{Select the Data}
\label{sec:select-data}
Selecting appropriate datasets is critical for effective multilingual hate speech detection and counterspeech generation, as they must capture linguistic diversity, cultural nuances, and platform-specific characteristics \citep{fortuna2018survey, chung2019conan}. High-quality datasets enable robust model training across diverse linguistic and cultural contexts, but challenges such as class imbalance, bias, data scarcity in low-resource languages, and translation fidelity persist. Hate speech datasets, often sourced from social media platforms like Twitter and Facebook, support binary or multi-label classification tasks, while counterspeech datasets provide hate speech-counter-narrative pairs or triplets for classification and generation tasks. Tables ~\ref{tab:hatespeech_datasets} and ~\ref{tab:counterspeech_datasets} summarize key datasets, detailing their languages, sizes, and sources. Researchers should prioritize datasets with robust annotations, diverse sources, and bias mitigation strategies, tailoring selections to 
target languages and tasks.
\subsection{Multilingual Hate Speech Datasets}
\label{app:hate_speech_datasets}

High-quality datasets are essential for multilingual hate speech detection, enabling models to address diverse linguistic and cultural contexts \citep{fortuna2018survey}. Typically sourced from social media platforms like Twitter and Facebook, these datasets are annotated for binary (hate vs. non-hate) or multi-label tasks, but often face challenges such as class imbalance, bias, and data scarcity in low-resource languages. This subsection categorizes datasets by language groups, highlighting their characteristics and considerations for selection to inform effective task design.

\subsubsection{Indo-European Language Datasets}
Datasets for Indo-European languages, such as English, Spanish, German, and Italian, are widely available, often sourced from Twitter due to its accessibility. \citet{ousidhoum2019multilingual} provide a dataset of English, French, and Arabic tweets, annotated for multi-aspect hate speech (e.g., directness, target group), suitable for fine-grained classification but limited by its small size. \citet{huang2020multilingual} present a multilingual Twitter corpus across English, Italian, Polish, Portuguese, and Spanish, augmented with inferred demographic attributes (e.g., age, gender, race), enabling fairness evaluation but requiring validation of inferred labels. \citet{gertner2019mitre, bojkovsky2019stufiit} utilize the HatEval dataset from SemEval-2019 Task 5 \citep{basile2019semeval}, comprising English and Spanish tweets targeting immigrants and women, with balanced hate/non-hate labels, ideal for binary classification but sensitive to feature distribution shifts. \citet{montariol2022multilingual} recombine HatEval \citep{basile2019semeval}, AMI \citep{fersini2020ami}, and HaSpeeDe \citep{bosco2018overview} datasets (English, Spanish, Italian) with new splits, ensuring comparable sizes for cross-lingual tasks, supplemented by sentiment and Named Entity Recognition (NER) datasets for auxiliary training. \citet{mnassri2024multilingual, mishra2021exploring} leverage the HASOC 2019 \citep{mandl2019overview} dataset (English, Hindi, German) from Twitter and Facebook, supporting binary and multi-aspect tasks (e.g., hate, offensive, profane), though skewed label distributions pose challenges. \citet{kotarcic2022human} introduce the Swiss Hate Speech Corpus, with over 422,000 comments from Swiss online newspapers in German, French, and dialects, annotated via a human-in-the-loop pipeline for binary and multi-label tasks, notable for its scale but imbalanced due to the natural prevalence of hate speech. \citet{bigoulaeva2021cross} use the English Stormfront \citep{de2018hate} dataset (10,000 forum posts) and GermEval \citep{wiegand2018overview} German tweets, simplified to binary labels, highlighting data scarcity in German. \citet{hashmi2025metalinguist} contribute a novel English and Norwegian Twitter dataset, annotated using Llama 3 for neutral/hateful labels, addressing low-resource Norwegian contexts. \citet{sohn2019mc} employ the Spanish HatEval \citep{basile2019semeval}, GermEval \citep{wiegand2018overview} 2018 (German), and HaSpeeDe 2018 \citep{bosco2018overview} (Italian) datasets, using translations to augment data, suitable for cross-lingual tasks but requiring caution for translation quality. \citet{trager2025mftcxplain} present MFTCXplain, a 3,000-tweet dataset (704 English, 621 Italian, 608 Persian, 1,067 Portuguese) with expert annotations for hate speech and moral categories, targeting underrepresented languages. \citet{ahmad2025ua} introduce UA-HSD-2025, a manually annotated dataset of 5,240 Arabic and Urdu tweets from X, with preprocessing and 85\% Cohen’s Kappa IAA, targeting low-resource hate speech data.

\subsubsection{Indian Language Datasets}
Indian languages, often low-resource, present challenges due to linguistic diversity and code-mixing. \citet{thapa2025natural} provide a curated dataset for Devanagari-script languages (Hindi, Nepali) from NEHATE \citep{thapa2023nehate}, NAET \citep{rauniyar2023multi}, IEHate \citep{jafri2023uncovering}, and CHUNAV \citep{jafri2024chunav}, annotated for binary hate speech and target identification (community, individual, organization), critical for low-resource settings. \citet{yadav2023hate} consolidate three Hinglish datasets (\citet{bohra2018dataset}, \citet{kumar2018aggression}, HASOC 2021 \citep{mandl2021overview}) into 20,600 instances, achieving near-balanced binary labels, ideal for code-mixed tasks but requiring preprocessing for social media noise. \citet{biradar2021hate} use the \citet{bohra2018dataset} dataset of 4,575 Hinglish tweets, slightly imbalanced, highlighting challenges with non-standard writing. \citet{shukla2024multilingual} leverage CONSTRAINT 2021 \citep{bhardwaj2020hostility} and HHSD \citep{kapil2023hhsd} for Hinglish, supporting multimodal inputs (text, images, videos), though dataset size details are limited. \citet{ghosh2025hate} employ HASOC datasets (Hindi, Marathi, Bangla) and newly annotated HS-Assamese and HS-Bodo datasets from social media, addressing under-resourced Northeast Indian languages with binary labels, though annotation costs are high. \citet{singh2024generalizable} introduce multicomb, a 300,000-text dataset across 12 Indian languages (e.g., Hindi, Bengali, Marathi) and English, sourced from ShareChat, YouTube, and Twitter, tackling code-mixing and cultural nuances but facing annotation inconsistencies. \citet{kodali2025bytesizedllm} use CHIPSAL@COLING 2025 datasets for Hindi and Nepali (19,019 train samples for detection, 2,214 for target identification), supplemented by Bhojpuri, Sanskrit, and Marathi for MLM fine-tuning, addressing Devanagari-script data scarcity.

\subsubsection{Low-Resource and Other Language Datasets}
Datasets for low-resource languages are vital for equitable detection. \citet{de2018multilingual} analyze eight corpora covering jihadist, extremist, racist, and sexist content in multiple languages, with the JIHADISM corpus (tweets) distinguishing hate/safe labels using cues like "kuffar," though manual annotation limits scalability. \citet{pacaldo2025leveraging} curate a secondary dataset from social media (Facebook, Twitter) in Cebuano, Tagalog, and English (88,000 post-cleaning instances, balanced to 176,000 via SMOTE), addressing low-resource scarcity with manual validation and preprocessing (tokenization, stemming). \citet{usman2025large} introduce a trilingual dataset of 10,193 annotated tweets from X (English: 3,834; Spanish: 3,162; Urdu: 3,197), with native annotators achieving 0.821 Fleiss’ Kappa, focusing on low-resource Urdu challenges like code-mixing.\citet{al2025multilingual} merge the L-HSAB \citep{mulki2019hsab} Arabic dataset (5,846 records) and an English dataset (24,783 records) into 30,629 records, addressing MENA region code-switching with SMOTE for imbalance, but requiring careful preprocessing. \citet{chavinda2025dual} provide Sinhala (Facebook: 6,345, Twitter: 4,502) and Tamil (5,503) datasets from social media, near-balanced for Sinhala but imbalanced for Tamil, capturing informal expressions in low-resource contexts. \citet{kousar2024mlhs} combine datasets for nine languages (e.g., Arabic, French, Spanish) and newly collected Danish, Turkish, and Hindi Twitter data, enhancing diversity but facing annotation challenges. 

\subsubsection{Challenges in Dataset Selection}
Selecting multilingual hate speech datasets requires balancing size, diversity, and quality. Class imbalance, prevalent in datasets like the Swiss Hate Speech Corpus \citep{kotarcic2022human} and HASOC 2019 \citep{mnassri2024multilingual}, necessitates techniques like SMOTE \citep{al2025multilingual}. Bias from collection methods, such as keyword-based sampling, can lead to false positives, particularly in Arabic datasets \citep{ousidhoum2020comparative, de2018multilingual}. Code-mixing in Hinglish \citep{yadav2023hate, biradar2021hate} and Indian language diversity \citep{singh2024generalizable} require preprocessing for noise and non-standard text. Low-resource languages like Assamese, Bodo \citep{ghosh2025hate}, Sinhala, and Tamil \citep{chavinda2025dual} suffer from data scarcity, necessitating cross-lingual or federated learning approaches \citep{singh2024generalizable}. Annotation inconsistencies across merged datasets \citep{singh2024generalizable} and cultural nuances (e.g., sarcasm in Swiss data \citep{kotarcic2022human}) complicate model training. Researchers should prioritize datasets with robust annotations, diverse sources, and bias mitigation strategies, tailoring selections to target languages and tasks.

\subsection{Multilingual Counterspeech Datasets}
\label{app:counterspeech_datasets}

Counterspeech datasets enable the development of constructive, culturally sensitive responses to combat online hate, requiring diverse voices and strategies \citep{chung2019conan}. These datasets provide hate speech-counter-narrative (HS-CN) pairs or triplets from social media, curated sources, or translated corpora. They vary in language coverage, size, and annotation methods, ranging from expert nichesourcing to AI-assisted annotation, while addressing challenges like cultural adaptation, translation fidelity, and data scarcity in low-resource settings. This subsection reviews six key datasets, detailing their characteristics and practical insights for designing impactful counterspeech tasks.

\subsubsection*{CONAN}
\textit{CONAN} \citep{chung2019conan} offers 15,024 HS-CN pairs across English (6,654), French (5,157), and Italian (3,213), with 4,078 original pairs expanded through paraphrasing and manual translation by non-experts. Sourced from social media and crafted by over 100 NGO operators targeting Islamophobia, its counter-narratives employ strategies like facts, humor, and warnings of consequences. Despite challenges with translation quality, its nichesourcing approach and human-evaluated quality make it a cornerstone for classification and generation tasks. Researchers can pair it with \textit{IndicCONAN} \citep{sahoo2024indicconan} for broader linguistic coverage or adopt its methodology to engage local experts for new datasets.

\subsubsection*{MultiTarget CONAN}
\textit{MultiTarget CONAN} \citep{fanton2021human} provides 5,000 English HS-CN pairs addressing diverse targets, including women, POC, and LGBT+. Built through a human-in-the-loop process with 880 seed pairs from NGO experts, it prioritizes informed, non-aggressive responses, though it lacks defined strategies. Evaluated for diversity and novelty via Jaccard similarity, it addresses multi-target coverage challenges.

\subsubsection*{ML-MTCONAN-KN}
\textit{ML-MTCONAN-KN} \citep{bonaldi2025first} delivers 2,384 triplets (hate speech, counterspeech, background knowledge) in Basque, English, Italian, and Spanish. Sourced from \textit{MT-CONAN} \citep{fanton2021human} and translated using DeepL and Itzuli with expert post-editing, it targets Jews, LGBT+, and immigrants. While lacking specific strategies, its knowledge integration supports nuanced generation tasks in low-resource languages like Basque. Evaluated with BLEU, ROUGE-L, and JudgeLM, it is ideal for experimenting with informed responses. Researchers can pair it with \textit{CONAN-EUS} \citep{bengoetxea2024basque} for Basque-focused projects or adopt its translation pipeline for new low-resource datasets.

\subsubsection*{CONAN-EUS}
\textit{CONAN-EUS} \citep{bengoetxea2024basque} provides 13,308 HS-CN pairs (6,654 each in Basque and Spanish), translated from English \textit{CONAN} \citep{chung2019conan} using Google API and refined by three native translators. Targeting Islamophobia with strategies like facts and hypocrisy, it ensures translation fidelity for cross-lingual generation. Evaluated for Relatedness and Coherence, it addresses data scarcity in Basque.

\subsubsection*{IndicCONAN}
\textit{IndicCONAN} \citep{sahoo2024indicconan} offers approximately 5,018 HS-CN pairs (2,509 each in Hindi and Indian English), crafted via human-in-the-loop with NLLB-200 and expert editing. Targeting religion, gender, and caste, its counter-narratives use denouncing and facts, addressing cultural relevance challenges. Evaluated with BLEU, METEOR, and BERTScore, it is ideal for code-mixed generation tasks. Researchers can scale its approach for other Indic languages, combine it with \textit{CONAN} \citep{chung2019conan} for global benchmarks, or use toxicity metrics to ensure response appropriateness.

\subsubsection*{Low-Resource Counterspeech Dataset}
\textit{Low-Resource Counterspeech} \citep{das2024low} provides 5,062 AS-CS pairs (2,460 Bengali, 2,602 Hindi) from Twitter, annotated by experts using strategies like empathy, humor, and consequences. Synthetic transfer from \textit{MultiTarget CONAN} \citep{fanton2021human} enhances its scope, though translation tools are unspecified. Evaluated with BLEU, ROUGE-1, and BERTScore, it tackles data scarcity for low-resource generation. Researchers can adopt its lexicon-based crawling for new datasets.

\begin{table*}[htbp]
\centering
\small
\begin{tabular}{@{}p{5cm}p{4cm}p{1.3cm}p{4cm}@{}}
\toprule
\textbf{Dataset} & \textbf{Languages} & \textbf{Size} & \textbf{Source/Method} \\
\midrule
Hindi-English Code-Mixed \citep{bohra2018dataset} & Hindi, English & 4,575 & Social media; code-mixed annotation \\
Amharic and Afaan Oromo Dataset \citep{ababu2025bilingual} & Amharic, Afaan Oromo & 30,000 & Facebook, Twitter; expert annotation \\
HatEval \citep{basile2019semeval} & English, Spanish & 19,600 & Twitter; keyword monitoring \\
English-Norwegian Dataset \citep{hashmi2025metalinguist} & English, Norwegian & 1,043 & Twitter; Llama 3 annotation \\
LAHM \citep{yadav2023lahm} & English, Hindi, French, Arabic, German, Spanish & ~300,000 & Twitter; semi-supervised annotation \\
AfriHate \citep{muhammad2025afrihate} & 15 African languages & 90,437 & Social media; native speaker annotation \\
Fine-Grained Multilingual Dataset \citep{siddiqui2024fine} & English, Urdu, Sindhi & ~47,000 & Translated MLMA, Cyberbully; annotated \\
\citet{ghosh2025hate} & Hindi, Marathi, Bangla, Assamese, Bodo & Not specified & Social media; native speaker annotation \\
Multi3Hate \citep{bui2024multi3hate} & English, German, Spanish, Hindi, Chinese & 300 memes & Social media; parallel memes \\
MHC \citep{rottger2022multilingual} & 10 languages & 36,582 & Handcrafted test cases; binary labels \\
Swiss Hate Speech \citep{kotarcic2022human} & German, French & 422,000+ & News comments; human-in-the-loop \\
Multicomb \citep{singh2024generalizable} & 12 Indian languages, English & ~300,000 & ShareChat, Twitter aggregation \\
Multilingual Twitter Corpus \citep{huang2020multilingual} & English, Italian, Polish, Portuguese, Spanish & 106,350 & Twitter; demographic info \\
Multi-Aspect \citep{ousidhoum2019multilingual} & English, French, Arabic & ~13,000 & Twitter; Amazon Turk annotation \\
Hinglish Consolidated \citep{yadav2023hate} & Hindi, English & 20,600 & Twitter; merged datasets \\
Arabic-English Unified \citep{al2025multilingual} & Arabic, English & 30,629 & Twitter; merged L-HSAB, mrmorj \\
HateCheckHIn \citep{das2022hatecheckhin} & Hindi & 5,884 & Handcrafted Roman/code-mixed Hindi \\
HASOC 2021 \citep{mandl2021overview} & English, Hindi & 6,126 & Twitter, Facebook; hate/offense labels \\
Devanagari Script \citep{thapa2025natural} & Hindi, Nepali & 74,889 & Social media; merged NEHATE, NAET \\
Philippine Multilingual HS \citep{pacaldo2025leveraging} & Cebuano, Tagalog, English & 176,000 (balanced) & Secondary social media \\
Trilingual HS (English-Spanish-Urdu) \citep{usman2025large} & English, Spanish, Urdu & 10,193 & Twitter \\
MFTCXplain \citep{trager2025mftcxplain} & English, Italian, Persian, Portuguese & 3,000 & Twitter benchmarks \\
UA-HSD-2025 \citep{ahmad2025ua} & Arabic, Urdu & 5,240 & Twitter \\
CHIPSAL@COLING 2025 \citep{kodali2025bytesizedllm} & Hindi, Nepali, Bhojpuri, Sanskrit, Marathi & 19,019 (Train B), 2,214 (Train C) & Shared task, curated \\
\bottomrule
\end{tabular}
\caption{Multilingual Hate Speech Datasets}
\label{tab:hatespeech_datasets}
\end{table*}

\begin{table*}[htbp]
\centering
\small
\begin{tabular}{@{}p{5cm}p{4cm}p{1.3cm}p{4cm}@{}}
\toprule
\textbf{Dataset} & \textbf{Languages} & \textbf{Size} & \textbf{Source/Method} \\
\midrule
CONAN \citep{chung2019conan} & English, French, Italian & 15,024 pairs & Social media; nichesourcing \\
MultiTarget CONAN \citep{fanton2021human} & English & 5,000 pairs & Social media; human-in-the-loop \\
ML-MTCONAN-KN \citep{bonaldi2025first} & Basque, English, Italian, Spanish & 2,384 triplets & MT-CONAN; translation \\
CONAN-EUS \citep{bengoetxea2024basque} & Basque, Spanish & 13,308 pairs & CONAN; translation \\
IndicCONAN \citep{sahoo2024indicconan} & Hindi, Indian English & 5,018 pairs & Social media; human-in-the-loop \\
Low-Resource \citep{das2024low} & Bengali, Hindi & 5,062 pairs & Twitter; crawling and annotation \\
\bottomrule
\end{tabular}
\caption{Multilingual Counterspeech Datasets}
\label{tab:counterspeech_datasets}
\end{table*}

\section{Evaluation}
\label{sec:evaluate}

Evaluating multilingual hate speech detection and counterspeech generation systems demands metrics and methods that not only measure performance across diverse languages but also ensure cultural fairness and inclusivity. Standard evaluation pipelines designed for English-centric tasks often fall short when applied to multilingual or code-mixed contexts, where linguistic diversity, dialectal variation, and socio-cultural specificity play critical roles in interpretation \citep{ousidhoum2020comparative, rottger2022multilingual}. For hate speech detection, this means that a model achieving high accuracy in English may underperform in low-resource languages due to imbalanced datasets or contextually inappropriate translations \citep{chan2024hate}. Similarly, counterspeech evaluation presents unique challenges: unlike classification tasks, which can rely on precision and recall, counterspeech requires assessing qualities such as persuasiveness, empathy, non-toxicity, and cultural appropriateness, which are difficult to capture with surface-level metrics \citep{das2024low, bonaldi2025first}.  

Multilingual evaluation also suffers from the \textit{translation fidelity problem}. Back-translation and automatic alignment methods are commonly used for benchmarking across languages, but these approaches risk losing cultural nuance and implicitly embedding majority-language bias \citep{huang2020multilingual, aluru2020deep}. In practice, harmful expressions such as sarcasm, coded speech, or community-specific slurs may be mistranslated into benign content, leading to inflated scores that misrepresent true system effectiveness \citep{lam2022end}. This problem is further compounded when systems are deployed at scale, where errors disproportionately affect marginalized communities.  

To address these issues, evaluation in multilingual hate speech and counterspeech research typically combines three complementary dimensions. First, \textit{quantitative metrics}—including precision, recall, F1-score, macro-averages, and multilingual extensions of fairness metrics—provide a scalable baseline for benchmarking, though they rarely capture cultural sensitivity \citep{huang2020multilingual}. Second, \textit{qualitative evaluation} via human judgment remains essential, enabling the measurement of persuasiveness, empathy, or harm reduction. Studies increasingly use expert annotators or community members to validate system outputs, though challenges remain in terms of cost, consistency, and inter-annotator agreement \citep{fanton2021human, sahoo2024indicconan}. Finally, \textit{best practices} are emerging that integrate hybrid evaluation strategies: combining automatic metrics with human-in-the-loop protocols, aligning quantitative performance with ethical and cultural considerations, and employing tools such as LLM-based judges (e.g., JudgeLM) for scalable cross-lingual evaluation \citep{moscato2025mnlp, bennie2025codeofconduct}.  

In this section, we critically assess these approaches, examining their strengths and limitations in both detection and counterspeech generation tasks. We emphasize that robust evaluation in multilingual contexts requires moving beyond traditional metrics toward frameworks that explicitly account for fairness, transparency, and cultural grounding, ensuring that systems perform reliably across diverse languages and communities.

\subsection{Quantitative Metrics}
\label{subsec:quantitative_metrics}
Quantitative metrics provide a foundation for assessing system performance, tailored to task type. For hate speech detection, classification metrics—accuracy, precision, recall, and F1-score—are standard. \citet{basile2019semeval} use macro F1-score in HatEval to balance performance across English and Spanish, addressing imbalanced classes. \citet{yadav2023lahm} employ F1-score and accuracy for six languages, while \citet{rottger2022multilingual} and \citet{das2022hatecheckhin} use accuracy on diagnostic test cases to reveal weaknesses in 10 languages and Hinglish, respectively. \citet{chavinda2025dual} achieve high F1-score for Sinhala and Tamil, outperforming CNN baselines. \citet{mishra2021exploring} apply macro F1-score for HASOC 2019 across English, Hindi, and German, assessing multi-task performance. \citet{yadav2023hate} report 0.876 accuracy and 0.835 F1-score for Hinglish, and \citet{shukla2024multilingual} achieve 0.88 accuracy for multimodal Hinglish inputs. \citet{al2025multilingual} use SMOTE-enhanced metrics for English-Arabic code-switching, while \citet{biradar2021hate} report 72\% accuracy for Hinglish TIF-DNN. \citet{mnassri2024multilingual} note a 9.23\% F1-score improvement with SS-GAN-mBERT, addressing data scarcity. Fairness metrics, like FNED/FPED \citep{huang2020multilingual}, and bias metrics (B1, B2) \citep{ousidhoum2020comparative}, enhance equity across five and seven languages, respectively. \citet{pacaldo2025leveraging} report mBERT's 96.1\% accuracy and 0.97 F1-score on combined Cebuano-Tagalog-English data, with language-specific AUC-ROC highlighting recall issues in low-resource Tagalog.

Cross-lingual evaluations add complexity. \citet{bigoulaeva2021cross} and \citet{montariol2022multilingual} use macro F1-score for zero-shot English-to-German and English-Spanish-Italian transfers, respectively, while \citet{hashmi2025metalinguist} report 79\% (zero-shot) and 90\% (few-shot) F1-scores for Norwegian-English. \citet{de2022unsupervised} and \citet{thapa2025natural} apply F1-score for unsupervised and Devanagari-script tasks, noting lower hate speech scores. \citet{mahajan2024ensmulhatecyb}, \citet{gertner2019mitre}, \citet{bojkovsky2019stufiit}, \citet{sohn2019mc}, and \citet{singh2024generalizable} emphasize macro F1-score and weighted accuracy for diverse languages, highlighting scalability but masking low-resource language disparities. These findings reinforce that while macro-averaged metrics reduce dominance of majority-class performance, they may still obscure errors in minority or code-switched varieties, calling for more granular per-class and per-language reporting.

For counterspeech generation, text quality metrics are critical. \citet{sahoo2024indicconan} use BLEU, ROUGE-L, BERTScore, and Self-BLEU for Hindi and Indian English, ensuring diversity. \citet{bonaldi2025first} apply BLEU, ROUGE-L, and JudgeLM for Basque, English, Italian, and Spanish, prioritizing contextual appropriateness. \citet{das2024low} use BLEU, METEOR, and ROUGE-1 for Bengali and Hindi, with BERTScore correlating strongly with human judgments. \citet{fanton2021human} measure diversity via Jaccard similarity, and \citet{sahoo2024indicconan} use toxicity scores to ensure safe outputs. However, BLEU may undervalue creative responses, as \citet{bonaldi2025first} note, demanding hybrid approaches. Recent work shows promise in integrating automatic and human-centered evaluation, for example combining JudgeLM with human ratings \citep{bennie2025codeofconduct}, or supplementing surface-form overlap scores with measures of empathy, persuasiveness, and cultural appropriateness \citep{muhammad2025afrihate}. These approaches highlight the importance of moving beyond reference-based overlap to richer, multidimensional evaluation frameworks. 

Overall, evaluation in multilingual hate speech and counterspeech tasks must balance \emph{scalability} with \emph{cultural fidelity}. While quantitative metrics offer reproducibility, they often underrepresent socio-cultural nuance. Hybrid pipelines—automated metrics for efficiency, human-in-the-loop judgments for sensitivity, and fairness metrics for equity—emerge as best practices.

\subsection{Qualitative Approaches}
\label{subsec:qualitative_approaches}

While quantitative metrics provide reproducibility and scalability, qualitative evaluation exposes dimensions that numbers alone cannot capture. It is particularly important in multilingual contexts, where meaning is shaped by cultural subtleties, code-switching practices, and socio-political framing. In such cases, even a high F1-score can obscure systematic errors if a model consistently misclassifies community-specific slurs, sarcasm, or honorifics.

For hate speech detection, qualitative assessment often begins with annotation practices. \textit{AfriHate} \citep{muhammad2025afrihate} illustrate this by relying on native speakers across 15 African languages, ensuring that definitions of hate speech align with local cultural norms. This stands in contrast to \textit{HatEval} \citep{basile2019semeval}, which blended crowd-sourced and expert annotations to balance coverage with reliability. Such differences highlight a persistent trade-off: broad annotation efforts can improve dataset scale, but without cultural expertise, they risk flattening nuanced categories into generic labels. Other studies emphasize transparency rather than scale. For example, \citet{siddiqui2024fine} apply LIME to visualize model reasoning for Urdu and Sindhi, offering a qualitative layer that allows annotators to interrogate why a system reached a decision. By contrast, \textit{Multi3Hate} \citep{bui2024multi3hate} reveal how cultural disagreement itself complicates evaluation, as annotators from the USA and India agreed on only 67\% of cases. These examples show that qualitative work is not merely an “add-on” to metrics but a lens into cultural contestation and interpretive diversity.

Counterspeech evaluation raises even more challenges. Unlike detection, where the outcome is binary or categorical, counterspeech quality depends on fluid notions of appropriateness, empathy, and persuasiveness. \citet{chung2019conan} demonstrate that expert annotators can achieve high agreement (Cohen’s Kappa = 0.92), yet agreement does not automatically equate to social effectiveness. A counter-response may be rated “appropriate” by experts while failing to resonate with the targeted community. Studies such as \citet{das2024low} and \citet{sahoo2024indicconan} acknowledge this gap by introducing dimensions like Suitableness, Specificity, and cultural appropriateness, bringing evaluation closer to real-world expectations. Yet the reliance on human raters also introduces subjectivity and potential bias: annotators may favor polite or fact-based responses, even though humor or irony might prove more persuasive in certain contexts. Moreover, datasets like \textit{CONAN-EUS} \citep{bengoetxea2024basque} and \textit{ML-MTCONAN-KN} \citep{bonaldi2025first} reveal that translation into low-resource languages complicates qualitative judgment—what seems coherent in English may sound awkward or even offensive when back-translated into Basque or Italian.

Taken together, qualitative approaches reveal a central paradox: they are indispensable for capturing cultural nuance, but they are also inherently fragile, relying on subjective judgments, variable cultural norms, and resource-intensive annotation pipelines. This suggests that the goal is not to replace quantitative metrics but to orchestrate a dialogue between the two. In practice, this means using human judgment not only to validate outputs but also to probe failure cases, refine taxonomies, and question whether the task definitions themselves align with community values. A promising direction lies in mixed-method evaluation—combining annotator insights, model explainability tools, and community feedback loops—to move from measuring “what the system outputs” toward evaluating “how the system impacts discourse.” In multilingual hate speech and counterspeech research, this shift from static scoring to socially grounded assessment may ultimately determine whether these systems achieve ethical relevance beyond benchmark datasets.

\textbf{Best Practices: } To address multilingual challenges, effective evaluation integrates quantitative and qualitative methods for robust, fair assessments. As outlined in Table~\ref{tab:best-practices_eval}, several complementary practices have emerged. Hybrid metrics that combine automated scores (e.g., macro F1 or BLEU) with human ratings balance technical accuracy with contextual appropriateness, though they demand higher resource investment. Ensuring cultural relevance through the involvement of native speakers enhances annotation fidelity across diverse contexts, while fairness and transparency are promoted by integrating bias-sensitive metrics (e.g., FNED/FPED) and explainability tools such as LIME. Standardized tasks, like HatEval, support cross-lingual comparability, but risk overlooking platform-specific or community-specific variations. Finally, dynamic updates to evaluation sets help models remain aligned with the evolving nature of hate speech and counterspeech online, though they require sustained data curation efforts. Collectively, these practices underscore that effective multilingual evaluation is not a one-time task but an ongoing, adaptive process that balances scalability, fairness, and cultural grounding (see Table~\ref{tab:best-practices_eval} for details)

\begin{table*}[t]
\centering

\resizebox{.8\textwidth}{!}{
\begin{tabular}{p{3cm}p{10cm}}
\toprule
 \textbf{Hybrid Metrics} & Combine macro F1-score with functional testing for hate speech \citep{rottger2022multilingual} and BLEU with human ratings for counterspeech \citep{das2024low, sahoo2024indicconan}. This ensures both technical accuracy and contextual relevance, though balancing automated and human inputs is resource-intensive. \\

\textbf{Cultural Relevance} &  Engage native speakers for annotations, as in \citet{muhammad2025afrihate, ababu2025bilingual, bengoetxea2024basque}, to capture linguistic nuances. This enhances evaluation accuracy in diverse contexts but requires expert coordination. \\

 \textbf{Fairness and Transparency} & Use FNED/FPED metrics \citep{huang2020multilingual} and XAI techniques like LIME \citep{siddiqui2024fine} to reduce bias and clarify model decisions. These promote equitable and interpretable systems, despite implementation complexity. \\
 
 \textbf{Standardized Tasks} &  Adopt shared tasks like HatEval \citep{basile2019semeval} for consistent evaluation across languages. This fosters comparability but may miss platform-specific variations. \\
 
 \textbf{Dynamic Updates} &  Update evaluation sets regularly \citep{paz2020hate} to reflect evolving hate speech and counterspeech patterns. This maintains system relevance but demands ongoing data curation. \\
\bottomrule
\end{tabular}}
\caption{Best Practices for Multilingual Evaluation}
\label{tab:best-practices_eval}
\end{table*}

\section{Open Challenges and Future Directions}
\label{sec:open_challenges}

Multilingual hate speech detection and counterspeech generation are crucial for combating online toxicity, but challenges like data scarcity, translation errors, and cultural biases hinder progress—especially in low-resource settings. Overcoming these requires innovative approaches to data, modeling, and evaluation.

\subsection{Challenges}
\label{subsec:challenges}

Despite advances in multilingual hate speech detection and counterspeech generation, several challenges persist. These obstacles are not isolated but deeply interconnected: data scarcity undermines model training, translation inaccuracies exacerbate linguistic complexity, and cultural biases complicate annotation and evaluation. Together, they limit scalability, fairness, and the real-world impact of current systems.

\textbf{Data Scarcity in Low-Resource Languages.} Limited annotated datasets for languages like Sinhala, Tamil \citep{chavinda2025dual}, Bengali, Hindi \citep{das2024low}, Basque \citep{bengoetxea2024basque}, and African languages \citep{muhammad2025afrihate} hinder model generalization. Manual annotation is costly, time-consuming, and unscalable, particularly when nuanced cultural context must be preserved. Machine-translated data offers a partial solution but often lacks quality, with errors that distort semantics and tone \citep{bengoetxea2024basque}. These issues disproportionately affect counterspeech generation, where even small inaccuracies can turn a constructive response into one that fails pragmatically or culturally. Without targeted investment in high-quality multilingual corpora, progress risks remaining concentrated in high-resource languages.

\textbf{Limited Multilingual Datasets.} Comprehensive multilingual corpora are scarce, restricting cross-lingual transfer and fairness evaluation. \citet{mnassri2025rag} mitigate this with a balanced Hate Speech Superset and Wikipedia integration, but the 1,000-test-sample limit per language constrains broader multilingual hate speech analysis. This lack of scale limits robust benchmarking and prevents fair comparisons across languages. In addition, demographic information is often missing: \citet{huang2020multilingual} note the absence of datasets with author demographics, which hinders bias assessment. Counterspeech faces an additional challenge: ephemerality. \citet{chung2019conan} highlight that counterspeech data disappears quickly due to content deletion, making it particularly difficult to sustain corpora in non-English languages such as Basque and Italian \citep{bonaldi2025first}. The volatility of data reduces reproducibility and undermines longitudinal analysis, leaving gaps in evaluation.

\textbf{Translation Inaccuracies and Linguistic Complexity.} Translation errors in parallel corpora degrade performance, especially for linguistically distant languages like Basque \citep{bengoetxea2024basque}. Code-mixing in Hinglish \citep{yadav2023hate} and Romanized text \citep{chavinda2025dual} further complicate feature extraction, while inconsistent annotation schemas make cross-lingual transfer difficult \citep{bigoulaeva2021cross}. \citet{ahmad2025ua} explore translation-based approaches (e.g., Arabic to Urdu via Google Translate), finding robust hate speech patterns but acknowledging inaccuracies in dialects, slang, and sarcasm. These limitations extend to counterspeech generation, where grammatical errors and awkward phrasing undermine persuasiveness in low-resource settings \citep{bengoetxea2024basque}. Task designers thus face a dilemma: translation pipelines enable rapid scaling but risk undermining authenticity and cultural appropriateness.

\textbf{Cultural Biases and Subjectivity.} Hate speech and counterspeech vary significantly across cultural and demographic contexts, introducing subjectivity at every stage of task design. \citet{bonaldi2024nlp} emphasize that implicit hate, such as stereotypes, is particularly challenging to detect consistently. \citet{muhammad2025afrihate} report cultural disagreements in African datasets, while \citet{bui2024multi3hate} identify multimodal annotation biases with low inter-annotator agreement (e.g., 67\% between annotators in the USA and India). These discrepancies reflect not just annotator inconsistency but deeper divergences in how communities interpret offensiveness and appropriateness. For Arabic and Urdu, \citet{ahmad2025ua} note difficulties in annotating sarcastic and disguised hate speech, which they partially mitigate with rigorous guidelines and high inter-annotator agreement (IAA). Yet subjectivity persists, especially for implicit forms. \citet{trager2025mftcxplain} address this by incorporating socio-political annotator metadata, showing how perspectives influence judgments. Even so, large language models struggle with moral sentiment in languages like Persian and Portuguese, leaving cultural bias gaps unresolved. These challenges highlight the tension between standardization and inclusivity: while uniform guidelines promote comparability, they may overlook culturally specific nuances essential to effective counterspeech.

\textbf{Evaluation Challenges for Counterspeech.} Assessing counterspeech impact is complex, as existing metrics do not adequately capture real-world effectiveness. BLEU and similar text similarity scores undervalue rhetorical success, missing whether a response actually reduces hostility \citep{hong2024outcome}. AI-generated counterspeech often appears overly polite or generic, diverging from natural human expression and limiting persuasive power \citep{song2024assessing}. Evaluating cultural specificity compounds the difficulty: a counterspeech strategy appropriate in one language community may fall flat or even backfire in another \citep{bengoetxea2024basque}. Proxy measures, such as conversation outcomes, offer partial insights but can introduce bias \citep{hong2024outcome}. Human evaluation remains the most reliable approach, as shown by \citet{furman2023high}, but it is costly, slow, and often inconsistent across annotators. This evaluation bottleneck hampers progress by obscuring whether improvements in model fluency translate into meaningful social impact.

\textbf{Synthesis.} These challenges are deeply interwoven. Data scarcity forces reliance on translation, which amplifies linguistic errors; limited multilingual datasets restrict fairness checks, allowing cultural biases to persist undetected; and the subjectivity inherent in annotation complicates both detection and evaluation. Counterspeech evaluation, in turn, is undermined by all of the above: poor data reduces diversity, translation errors distort tone, and cultural biases obscure judgments of effectiveness. The result is a fragile pipeline where weaknesses cascade from one stage to another. Addressing these issues requires holistic solutions that combine robust multilingual data collection, culturally informed annotation, improved translation handling, and evaluation metrics that prioritize pragmatic and ethical impact alongside accuracy. Only then can multilingual hate speech detection and counterspeech systems achieve both technical reliability and societal relevance.

\subsection{Future Directions}
\label{subsec:future_directions}

Future research in multilingual hate speech detection and counterspeech must build on current progress while addressing persistent challenges of data scarcity, cultural bias, and evaluation gaps. Promising avenues include expanding datasets, incorporating multimodality, improving cross-lingual models, developing standardized benchmarks, and enhancing the capabilities of large language models (LLMs). Each of these directions reinforces the others: richer datasets enable better cross-lingual transfer, multimodality complements textual analysis, and robust benchmarks ensure that advances are evaluated fairly and transparently.

\textbf{Expanding Multilingual Datasets.} High-quality, diverse datasets are essential for both detection and counterspeech generation. \citet{huang2020multilingual} propose including demographic attributes to improve fairness assessment, while \citet{chung2019conan} advocate expanding counterspeech datasets to cover varied hate targets such as migrants and LGBT+ groups \citep{sahoo2024indicconan, das2024low}. Shared tasks like ML-MTCONAN-KN \citep{bonaldi2025first} can play a key role by curating non-English data and encouraging community-wide participation. \citet{kodali2025bytesizedllm} emphasize that larger XLM-RoBERTa variants, paired with diverse datasets in Devanagari-script languages, enhance robustness in multilingual detection. Expanding datasets not only increases linguistic coverage but also supports culturally grounded evaluation, reducing the risk of reproducing dominant-language biases.

\textbf{Integrating Multimodal Data.} Online hate often extends beyond text, incorporating images, memes, audio, and video. Incorporating multimodality can therefore strengthen both detection and counterspeech. \citet{narula2024comprehensive} suggest building multimodal Hindi datasets, while \citet{bui2024multi3hate} propose joint text-image models for hate speech detection. On the counterspeech side, multimodal prompts—such as pairing text with images or memes—can enhance contextual specificity and persuasiveness \citep{furman2023high}. Integrating multimodal data also opens opportunities for richer evaluation, allowing researchers to study not only the linguistic form but also the visual and cultural framing of counterspeech.

\textbf{Improving Cross-Lingual Models.} Advances in cross-lingual modeling are vital to scaling detection and counterspeech to low-resource languages. Meta-learning approaches \citep{hashmi2025metalinguist} and federated learning strategies \citep{singh2024generalizable} mitigate data scarcity by learning from distributed or limited data while preserving fairness. Transformer-based methods support zero-shot transfer \citep{bigoulaeva2021cross}, and semi-supervised GANs allow unlabeled data to be leveraged effectively \citep{mnassri2024multilingual}. For counterspeech, cross-lingual classification \citep{chung2021multilingual} and translation-enhanced generation \citep{bengoetxea2024basque} provide scalable pathways to adapt strategies across languages. These methods bridge gaps between high- and low-resource settings, but their success depends on careful calibration to avoid amplifying translation errors or cultural mismatches.

\textbf{Developing Standardized Benchmarks.} Progress requires consistent, transparent evaluation frameworks. \citet{ousidhoum2020comparative} propose bias metrics (B1, B2) to quantify fairness, while \citet{bonaldi2025first} highlight shared tasks as a way to standardize counterspeech evaluation. Techniques such as data augmentation and SMOTE address dataset imbalance \citep{ghosh2025hate}, but evaluation must also evolve to capture real-world outcomes. \citet{hong2024outcome} argue for outcome-based evaluation, measuring whether counterspeech reduces hostility rather than only linguistic similarity. Standardized benchmarks can therefore integrate multiple layers—fairness, balance, pragmatic effectiveness—to ensure that models are not just technically strong but socially responsible.

\textbf{Enhancing LLM Capabilities.} Finally, the growing role of LLMs presents opportunities for more adaptive and human-like counterspeech. Fine-tuning methods can improve cultural relevance and reduce overly generic outputs \citep{hengle2024intent, farhan2025hyderabadi}. Advanced prompting strategies \citep{saha2024zero} and Direct Preference Optimization \citep{wadhwa2024northeastern} align model outputs with human preferences, producing counterspeech that is more persuasive, context-sensitive, and ethically aligned. As LLMs become central to multilingual NLP, refining their capacity to handle low-resource and culturally diverse languages will be key to global deployment.

\textbf{Synthesis.} These directions are mutually reinforcing. Expanded datasets underpin multimodal and cross-lingual research, while improved benchmarks ensure that advances are measured against fairness and real-world impact. LLM fine-tuning connects directly to cultural sensitivity, and multimodality opens pathways for more naturalistic counterspeech interventions. Together, these priorities chart a roadmap for moving beyond technical optimization toward socially grounded, inclusive systems. Future work should actively involve diverse linguistic communities, ensuring that research agendas reflect global needs rather than only high-resource contexts. By centering fairness, cultural alignment, and inclusivity, the field can move closer to building systems that not only detect hate but also counter it effectively across languages and cultures.

\section{Ethical Considerations}
\label{sec:ethical_considerations}

The studies on the detection of multilingual hate speech and counterspeech entail substantial ethical issues needing diligent consideration to facilitate responsible research work and application. Ethical diligence is not a peripheral matter but central to ensuring that systems developed in this domain do not replicate existing inequalities or unintentionally exacerbate social tensions.

\subsection{Bias and Fairness}
Hate speech datasets such as those shown in Tables~\ref {tab:hatespeech_datasets} and ~\ref{tab:counterspeech_datasets} are typically sourced from social media platforms such as Twitter and are prone to data collection biases such as keyword-based sampling, which can over-sample certain hate categories (e.g., migrants) while under-representing others (e.g., caste-based hate in Indic datasets) \citep{ousidhoum2020comparative, sahoo2024indicconan}. These biases can result in models that exhibit unfair performance between communities and languages, especially low-resource languages such as Amharic and Basque \citep{muhammad2025afrihate, bengoetxea2024basque}. To address this issue, we suggest data curation with diverse datasets that include native speakers, as demonstrated by \citet{muhammad2025afrihate}, as well as the utilization of fairness metrics such as FNED and FPED to quantify demographic biases \citep{huang2020multilingual}.
Researchers are encouraged to employ methods such as SMOTE to correct class imbalances, thus providing fair model performance in underrepresented groups \citep{al2025multilingual}. Beyond dataset balancing, fairness audits should be routine in multilingual hate speech pipelines, with explicit reporting of demographic error rates and transparent disclosure of performance gaps across linguistic communities. Only through systematic fairness evaluation can these systems avoid perpetuating structural biases in online discourse.

\subsection{Cultural Awareness}
Both hate speech and counterspeech are inherently context-dependent and differ significantly across cultural and language lines \citep{bui2024multi3hate}. False positives in detection or potentially inappropriate counterspeech responses, which might exacerbate conflicts, may be caused by misunderstanding cultural nuances like sarcasm or humor \citep{bonaldi2024nlp}. For instance, translations of counterspeech corpora such as CONAN-EUS \citep{bengoetxea2024basque} might not convey cultural context adequately and therefore be less effective in low-resource languages. We emphasize the native speaker annotation and expert validation, as in \textit{IndicCONAN} \citep{sahoo2024indicconan}, to validate cultural suitability. Researchers must interact with locals to develop hate speech and counterspeech taxonomies, particularly in regions of unique social dynamics, such as African or South Asian settings \citep{muhammad2025afrihate, sahoo2024indicconan}. Importantly, cultural awareness is not static: societal norms shift over time, meaning that taxonomies, counterspeech strategies, and annotation guidelines require periodic updates in dialogue with affected communities.

\subsection{Privacy and Anonymity}
Hate speech datasets typically contain privacy-sensitive user-generated content from platforms like Twitter, Reddit, and Facebook, raising privacy concerns \citep{yadav2023lahm}. Anonymized datasets could even be de-anonymized based on demographic characteristics \citep{huang2020multilingual}. We support ethical data management in our survey by encouraging strict anonymization protocols and compliance with data privacy regulations, e.g., the General Data Protection Regulation (GDPR) for EU datasets. Federated learning approaches, as proposed by \citet{singh2024generalizable}, offer a privacy-preserving approach for low-resource Indian languages with model training without centralizing sensitive data. Researchers must thoroughly document data sources and anonymization methods to maintain user trust and ethical integrity. Furthermore, access to raw datasets should be controlled under responsible licensing to prevent misuse, ensuring that sensitive material is only available for legitimate academic and applied research purposes.

\subsection{Potential for Harm}
The use of offensive hate speech examples, such as in Tables~\ref{tab:HS-example} and \ref{tab:conan-example}, is required for research but can cause harm if not handled properly \citep{mathew2021hatexplain, chung2019conan}.
To mitigate harm, we recommend that researchers limit exposure to offensive content to that which is strictly necessary for analysis and use controlled environments to access datasets. Counterspeech generation is also risky, as poorly worded responses can end up reinforcing harmful narratives or appearing dismissive \citep{hong2024outcome}. To avoid this, we endorse human-in-the-loop validation, as in \citet{fanton2021human}, and toxicity scoring, as in \citet{sahoo2024indicconan}, in order to render counterspeech non-aggressive and culturally sensitive. An additional consideration is the psychological impact on annotators, developers, and moderators repeatedly exposed to hateful content. Institutions should provide mental health support, clear guidelines, and rotation schemes to minimize long-term exposure risks.

\subsection{Transparency and Accountability}
Transparency in model development and evaluation is crucial for accountability purposes, particularly when it comes to deploying systems in real-world settings. Explainable AI techniques, such as LIME \citep{siddiqui2024fine}, enhance model interpretability, enabling stakeholders to understand detection and counterspeech decisions. We also advocate for open disclosure of limitations, as described in Section~\ref{sec:limitations}, including limitations in low-resource language coverage and evaluation metric limitations. Researchers need to engage with cross-disciplinary stakeholders, including ethicists and community members, to test the societal impact of their systems and ensure compliance with ethical guidelines. Accountability should further extend to impact assessments and clear lines of responsibility for harms caused by automated interventions, ensuring that affected communities have avenues for redress when systems fail.

\subsection{Ethical Implementation}
The implementation of hate speech detection and counterspeech mechanisms in multilingual settings needs rigorous evaluation of their social implications. AI-driven mechanisms can end up silencing valid expressions or struggle to identify covert hate, especially in code-mixed languages or low-resource contexts \citep{yadav2023hate, chavinda2025dual}. Further, counterspeech mechanisms must refrain from producing responses that fuel tensions or are not culturally appropriate \citep{bengoetxea2024basque}. We recommend continuous monitoring with human-in-the-loop pipelines, as in \citet{kotarcic2022human}, and regular updates to adapt to evolving hate speech patterns \citep{paz2020hate}. Collaboration with platform moderators and policymakers can ensure systems meet legal and ethical standards in various jurisdictions. Ethical implementation also involves anticipating adversarial misuse, such as attackers generating “toxic counterspeech” to discredit legitimate voices. Building safeguards against such risks is essential to protecting the integrity of these systems. Through addressing these ethical issues, our survey seeks to inform researchers towards the creation of fair, culturally responsive, and responsible frameworks for multilingual hate speech identification and counterspeech generation. These are key principles that can be used to create safe and inclusive online environments globally.

\subsection{Toward Responsible Global Deployment}
The ethical challenges discussed above demonstrate that multilingual hate speech detection and counterspeech generation cannot be separated from their societal context. Addressing bias, cultural awareness, privacy, harm reduction, transparency, and implementation concerns requires a holistic approach where technical development and ethical reflection proceed in tandem. While each subsection highlights distinct issues, together they point toward a unified responsibility: ensuring that systems designed for online safety do not reinforce inequities or silence marginalized voices. Responsible global deployment should therefore be incremental, community-driven, and adaptable. Incremental deployment allows for careful monitoring of unintended effects before scaling across platforms and languages. Community-driven development foregrounds the voices of those most affected by hate speech, ensuring that taxonomies, counterspeech strategies, and evaluation frameworks reflect local realities rather than external assumptions. Finally, adaptability is essential: hate speech evolves with political and cultural shifts, and systems must be regularly updated to remain relevant, fair, and effective. By embedding these principles into research practice, scholars and practitioners can move beyond technical performance benchmarks toward genuine societal impact. The long-term vision is not simply the mitigation of online hate, but the promotion of inclusive digital spaces where freedom of expression is preserved alongside safety and dignity for all communities.
\section{Limitations}
\label{sec:limitations}

The focus and extent of this paper are guided by a number of conceptual and methodological limitations, which are important in establishing the boundaries of our approach and the wider discipline. These limitations reflect not only choices in scope but also structural constraints of current research, which together highlight the areas where caution, refinement, and future development are necessary.

First, the focus we put on text-based frameworks limits the survey's applicability in addressing the rising incidence of multimodal hate speech, which involves modes like memes, videos, or audio on platforms such as TikTok, Telegram, and Instagram. Although we cite datasets such as Multi3Hate \citep{bui2024multi3hate} that contain text-image pairs, our task design and evaluation frameworks (Sections~\ref{sec:Task-Design} and \ref{sec:evaluate}) are overwhelmingly text-based. This decision follows the maturity of NLP based on text but underestimates the challenge of visual or audio cues, which often amplify hate through subtle cultural references, coded imagery, or paralinguistic signals like tone and prosody. As a result, our survey risks underrepresenting contexts where multimodality is not peripheral but central to the way hate and counterspeech are enacted.

Second, the survey's focus on generation and classification tasks presumes that hate speech and counterspeech have standardized definitions that apply across contexts, which simplifies their subjective, evolving nature. Our taxonomies (Section~\ref{sec:Background}) draw on established systems like \citet{ousidhoum2019multilingual} and \citet{chung2019conan}, yet these frameworks may fail to capture emergent forms of implicit hate, such as dogwhistles or coded language circulating in niche online communities. The assumption of definitional stability risks overgeneralizing the usefulness of our roadmap, particularly on platforms where slang, memes, and subcultural discourse evolve rapidly. A more dynamic taxonomy that incorporates user intent, community norms, and platform-specific lexicons would enhance task robustness, but it requires ongoing community input and longitudinal monitoring, which our survey does not exhaustively provide.

Third, our evaluation framework (Section~\ref{sec:evaluate}) prioritizes scalability through widely used measures like F1-score and BLEU, but this comes at the cost of contextual effectiveness. For counterspeech, BLEU favors syntactic similarity to reference responses, potentially rewarding safe but generic replies while undervaluing creative or culturally resonant ones \citep{bonaldi2025first}. For hate speech detection, macro F1-score may mask disparities between languages, especially in low-resource settings such as Basque or Amharic, where small datasets introduce inflated variance. These compromises limit the practical value of our survey for high-stakes moderation contexts, where misclassifications can have serious real-world consequences. More context-aware evaluation, such as counterspeech engagement rates, downstream behavioral changes, or demographic-specific error patterns, would offer richer insights, but such metrics remain underexplored in current research and beyond the scope of our study.

Fourth, the survey’s dependence on existing datasets (Tables \ref{tab:hatespeech_datasets} and \ref{tab:counterspeech_datasets}) narrows its coverage of hate speech and counterspeech in less-studied environments, such as professional or educational contexts (e.g., LinkedIn or academic forums). The majority of datasets are derived from Twitter, YouTube, or other informal, high-traffic platforms \citep{yadav2023lahm, singh2024generalizable}, which means our framework is more representative of public, colloquial interactions than of institutional or professional discourse. This imbalance risks reinforcing assumptions about the form and visibility of hate, while neglecting subtler but equally damaging manifestations in formal spaces.

Finally, the scope of our survey is shaped by its reliance on published methods and benchmarks, which creates a degree of hindsight bias. Much of the work we review reflects what has been systematically annotated and studied, rather than the full spectrum of how hate and counterspeech manifest in practice. This leaves gaps around under-documented languages, emergent platforms, and ephemeral content that are not yet captured in research pipelines.

Taken together, these limitations highlight structural trade-offs in surveying an extensive, multilingual domain. A focus on text ensures analytic clarity but downplays multimodal complexity; reliance on established taxonomies supports comparability but risks rigidity; scalable evaluation frameworks facilitate benchmarking but fail to capture pragmatic impact; and existing datasets provide coverage but bias the field toward certain languages and contexts. Through critical examination of these constraints, we emphasize the need for adaptive, multimodal, and context-sensitive approaches that move beyond static definitions and standardized metrics. Addressing these limitations will require deeper engagement with linguistic communities, cross-disciplinary collaboration, and evaluation strategies that prioritize both inclusivity and real-world effectiveness.

\section{Conclusion}
\label{sec:conclusion}

This survey has provided a comprehensive overview of multilingual hate speech detection and counterspeech generation, highlighting both the technical progress and the ethical complexities of this research area. We reviewed the state of the art across task design, dataset creation, model development, evaluation strategies, and ethical considerations, paying particular attention to the challenges that arise in low-resource and culturally diverse contexts. Our analysis underscores that while the field has advanced considerably through multilingual NLP and cross-lingual transfer, persistent gaps remain in fairness, inclusivity, and cultural alignment. A recurring theme across this paper is that technological solutions alone are insufficient. Effective detection and counterspeech systems must be deeply informed by community engagement, native-speaker validation, and cross-disciplinary collaboration. This entails not only constructing robust models, but also ensuring that the data underpinning them is representative, that evaluation frameworks extend beyond surface-level metrics, and that ethical safeguards are integrated throughout the research pipeline. Looking forward, the sustainable development of multilingual hate speech and counterspeech systems depends on three intertwined commitments. First, technical innovation must continue to push the boundaries of cross-lingual learning, multimodal integration, and fairness-aware modeling. Second, ethical responsibility must remain central, with transparency, accountability, and harm reduction guiding deployment decisions. Third, global collaboration among researchers, practitioners, policymakers, and communities is essential to build systems that are not only effective, but also just, culturally responsive, and socially beneficial. Ultimately, the goal is not merely to detect and counter hate, but to foster safer, more inclusive digital environments that support democratic dialogue and respect the dignity of diverse voices worldwide. By integrating rigorous technical approaches with ethical reflection and community partnership, future work can move the field closer to this vision and contribute meaningfully to the creation of healthier online ecosystems.

\bibliographystyle{ACM-Reference-Format}
\bibliography{references} 

\end{document}